\documentclass{article}
\usepackage{amssymb}
\usepackage{lipsum}
\usepackage{fontawesome5}
\usepackage{nicematrix}
\usepackage{booktabs}
\usepackage{amssymb}
\usepackage{pifont}

\usepackage{makecell}
\usepackage{multirow}
\usepackage{svg}
\usepackage{graphicx}
\usepackage{caption}
\usepackage{wrapfig}
\usepackage{subcaption}

\usepackage{enumitem}
\usepackage{float}
\usepackage{xspace}

\usepackage{amsmath}
\newcommand{\narrate}{\textsc{Narrate2Nav }}
\usepackage[final]{corl_2025} 
\title{\Large Narrate2Nav: Real-Time Visual Navigation with Implicit Language Reasoning in Human-Centric Environments}
\definecolor{myyellow}{RGB}{255, 215, 0}
\author{
  Amirreza Payandeh$^{1}$ \quad
  Anuj Pokhrel$^{1}$ \quad
  Daeun Song$^{1}$ \quad
  Marcos Zampieri$^{2}$ \quad
  Xuesu Xiao$^{1}$ \\
  $^{1}$Department of Computer Science, George Mason University \\
  $^{2}$Department of Information Sciences and Technology, George Mason University
}

\begin{document}
\maketitle

\begin{abstract}

Large Vision-Language Models (VLMs) have demonstrated potential in enhancing mobile robot navigation in human-centric environments by understanding contextual cues, human intentions, and social dynamics while exhibiting reasoning capabilities. However, their computational complexity and limited sensitivity to continuous numerical data impede real-time performance and precise motion control. 
To this end, we propose \textsc{Narrate2Nav}, a novel real-time vision-action model that leverages a novel self-supervised learning framework based on the Barlow Twins redundancy reduction loss to embed implicit natural language reasoning, social cues, and human intentions within a visual encoder—enabling reasoning in the model’s latent space rather than token space. The model combines RGB inputs, motion commands, and textual signals of scene context during training to bridge from robot observations to low-level motion commands for short-horizon point-goal navigation during deployment. Extensive evaluation of \textsc{Narrate2Nav} across various challenging scenarios in both offline unseen dataset and real-world experiments demonstrates an overall improvement of 52.94\% and 41.67\%, respectively, over the next best baseline. Additionally, qualitative comparative analysis of \textsc{Narrate2Nav}'s visual encoder attention map against four other baselines demonstrates enhanced attention to navigation-critical scene elements, underscoring its effectiveness in human-centric navigation tasks. 

\begin{figure*}[!htbp]
    \centering
    \includegraphics[width=\textwidth]{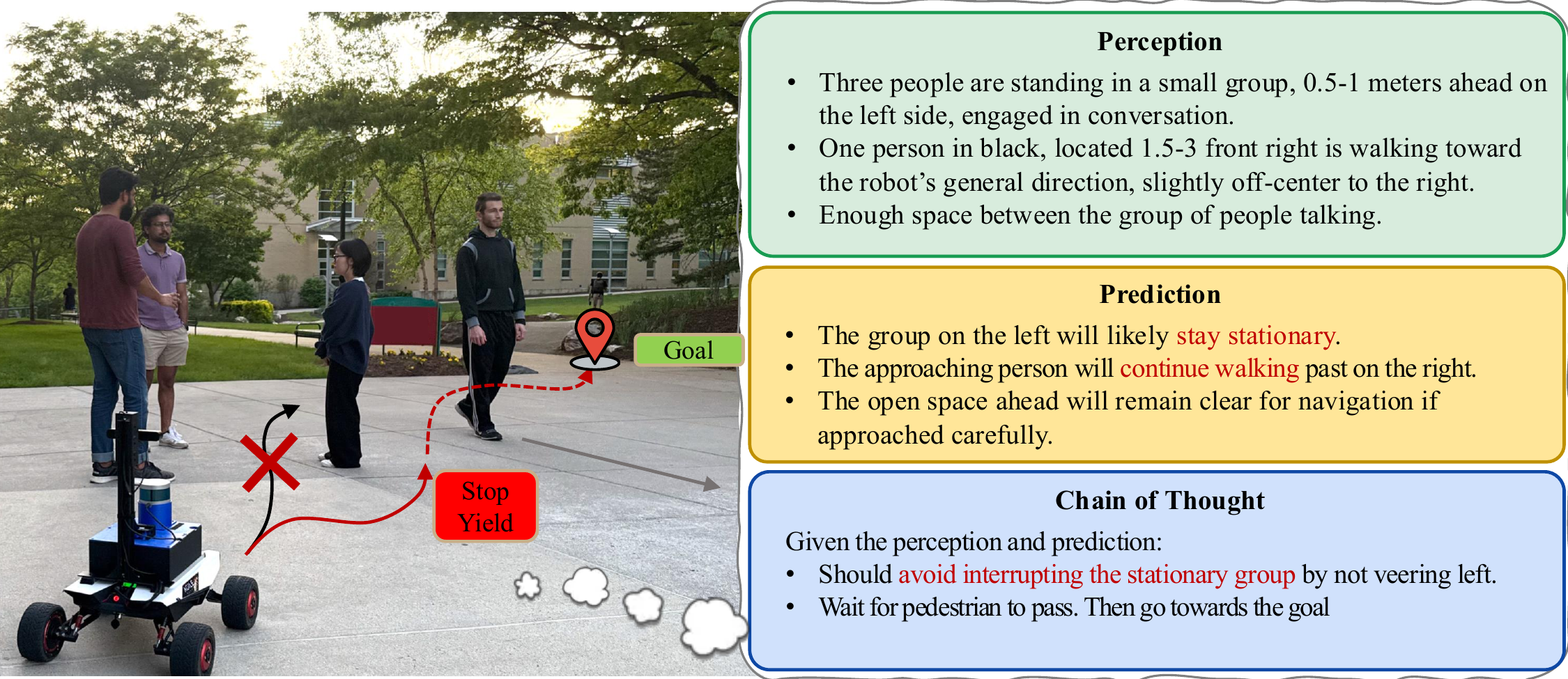}
    \caption{\textsc{Narrate2Nav}: Implicit natural language reasoning to navigate in human-centered environments while understanding the social cues and contextual information.}
    \label{fig:narrate2nav-abstract}
\end{figure*}
\end{abstract}

\section{INTRODUCTION}
\label{sec::introduction}

A delivery robot moves through a festival, navigating a dynamic environment where a group of children crosses its path chasing a drifting balloon, while a couple ahead stops unexpectedly to take a selfie, and a few people are chatting with enough space between them for the robot to pass. Instead of colliding with the children, freezing in place, or using the path between the people having a conversation, the robot smoothly adjusts its path, yielding to the children and navigating around the photo-taking couple and the chatting group. This scenario highlights the need to enhance the navigation stack with higher-level reasoning capabilities that adhere to complex and subtle social rules, such as understanding why a human is moving in a certain way, when to yield, and who will cross the path. To address this, robots need to accurately predict human intentions, interpret social cues, and employ human-like reasoning to take socially aware actions.

Traditional methods often rely on rule-based systems or predefined path-planning algorithms, which struggle to adapt to dynamic, human-centric environments and typically lack the flexibility to incorporate social norms or interpret subtle contextual cues~\cite{francis2025principles, mirsky2024conflict, mavrogiannis2023core}. Learning-based approaches, such as imitation learning and reinforcement learning~\cite{xiao2022motion,10323465,9679193}, which learn from demonstration~\cite{musohu,dtg, rethinking} or trial and error~\cite{kretzschmar16ijrr}, have shown promising results. However, these methods require collecting substantial amount of data. Additionally, they often fail to understand social cues and environmental context and do not achieve human-like reasoning through pixel values.

With recent advancements in large Vision-Language Models (VLMs) demonstrating an understanding of contextual information, human intentions, and social cues, several research efforts have explored the integration of natural language to enhance robots’ navigation performance using human-like reasoning~\cite{vlm-social,social-llava, olivianav,susceptible, song2025vl}. 
However, current approaches rely on large VLMs, whose heavy computational demands restrict their applicability in real-time scenarios. Additionally, due to their insensitivity to continuous numerical data, many existing VLM-based navigation frameworks prompt VLMs to generate high-level linguistic macro-actions, which are difficult to ground into real-world, low-level motion commands~\cite{vlm-social}.
To this end, we ask the question:
\emph{How can we leverage human-like language reasoning in mobile robot navigation while meeting real-time constraints?}

In this work, we present \narrate a novel real-time vision-action model that implicitly integrates human-like language reasoning to bridge robot visual observations and low-level motion commands for short-horizon point-goal navigation in dynamic, crowded environments. Our approach embeds language-based reasoning, social cues, and contextual awareness into a visual encoder through a novel pre-training method using the Barlow Twins redundancy reduction loss~\cite{barlow}. Specifically, \narrate enriches the current state representation with a multi-modal future state embedding that incorporates RGB visual inputs, low-level motion commands, and textual descriptions such as scene context, human intentions, trajectory summaries, and chain-of-thought (CoT) reasoning. This enables the model to explicitly leverage multi-modal signals during training, while relying only on RGB inputs with implicit language reasoning occurring in latent space rather than token space during navigation inference.

The summary of our contributions is as follows:
\vspace{-0.5em}
\begin{itemize}
\item[1.]\textsc{Narrate2Nav}, a novel real-time vision-action model baseline that bridges the gap between visual observations and low-level motion commands, implicitly considering human-like language reasoning in human-centric environments.
\item[2.] A new Self-Supervised Learning (SSL) framework that refines the visual encoder's attention maps using textual signals to focus on image regions relevant to the navigation task and enhance its ability to operate effectively in dynamic environments.

\item[3.] Extensive analysis of state-of-the-art learning-based mobile robot navigation models across various challenging scenarios in human-centered environments, including both real-world experiments and offline-based evaluations, demonstrating a 52.94\% and 41.67\% improvement of \textsc{Narrate2Nav} over the next best baseline, respectively.

\end{itemize}

\section{RELATED WORK}
\label{sec::related_work}
\paragraph{Visual Navigation:}
GNM~\cite{gnm} combined various datasets with different robot embodiments to learn an omni-policy for visual navigation, mapping RGB inputs to low-level motion commands. ViNT~\cite{vint} replaced the GNM model by introducing a Transformer-based, end-to-end behavioral cloning model trained on diverse real-world robot datasets to enable zero-shot generalization across different robot embodiments. NoMAD~\cite{nomad} introduced a unified diffusion policy that supported both goal-conditioned navigation and task-agnostic exploration through goal masking. CityWalker~\cite{citywalker} proposed a pipeline to learn a navigation policy in dynamic urban environments using in-the-wild city walking video data. However, all these approaches focus solely on mapping visual observations to actions and fall short of embedding social context, human intentions, and human-like reasoning essential for navigation in dynamic, human-centric environments, which is the gap \textsc{Narrate2Nav} fills. 
\vspace{-0.5em}
\paragraph{Representation Learning for Navigation:}
Eftekhar et al.~\cite{selective} introduced a task-conditioned bottleneckusing a small learnable codebook module to selectively filter point of interest in visual observations. VANP~\cite{vanp} and Vi-LAD~\cite{vilad} refined visual attention maps using action signals. CAHSOR~\cite{cahsor} explored human preference learning and competence awareness for off-road navigation through the use of SSL. ViPlanner~\cite{viplanner} introduced a semantic-aware, end-to-end local path planning framework trained entirely in simulation by fusing semantic segmentation with depth information to better assess terrain traversability and enable zero-shot transfer to real-world environments. NavFormer~\cite{navformer} presented a transformer-based policy with dual visual encoders for target-driven navigation in unfamiliar, dynamic settings, trained through cross-task learning for exploration and collision avoidance. 
\textsc{Narrate2Nav} is based on the hypothesis that textual descriptions of navigation scenarios are another useful modality to inform navigation decisions, which  none of the mentioned approaches utilized to train the visual encoder.

\vspace{-0.5em}
\paragraph{Language for Navigation:}
LeLaN~\cite{lelan} provides a dataset with language annotations derived from human walking and YouTube videos for the task of language-based object-goal navigation. In contrast, \narrate uses textual descriptions as a representation learning signal rather than as explicit navigation goals.
RING~\cite{onering} trained a transformer-based model on large-scale simulation data for generalization to unseen embodiments for object goal navigation. Social-LLaVA~\cite{social-llava} proposed using natural language to bring human-like reasoning to mobile robot navigation in crowds, generating high-level navigation actions through CoT reasoning expressed in natural language. OLiVia-Nav~\cite{olivianav} is the closest work to \narrate and used GPT-4o to label the MuSoHu~\cite{musohu} dataset with textual social context for social robot navigation using a LiDAR and an RGB camera. In contrast, to the best of our knowledge, \narrate is the first of its kind to embed 3D structural information through weak supervision of textual signals into 2D images for RGB-only visual navigation. Moreover, \narrate is the first to apply CoT reasoning for training a vision-action model for visual navigation in human-centric environments.


\section{Narrate2Nav}
\label{sec::Narrate2Nav}
\setlength{\abovecaptionskip}{2pt}  
\setlength{\belowcaptionskip}{2pt}  
\begin{figure*}[tbp]
    \centering
    \includegraphics[width=\textwidth]{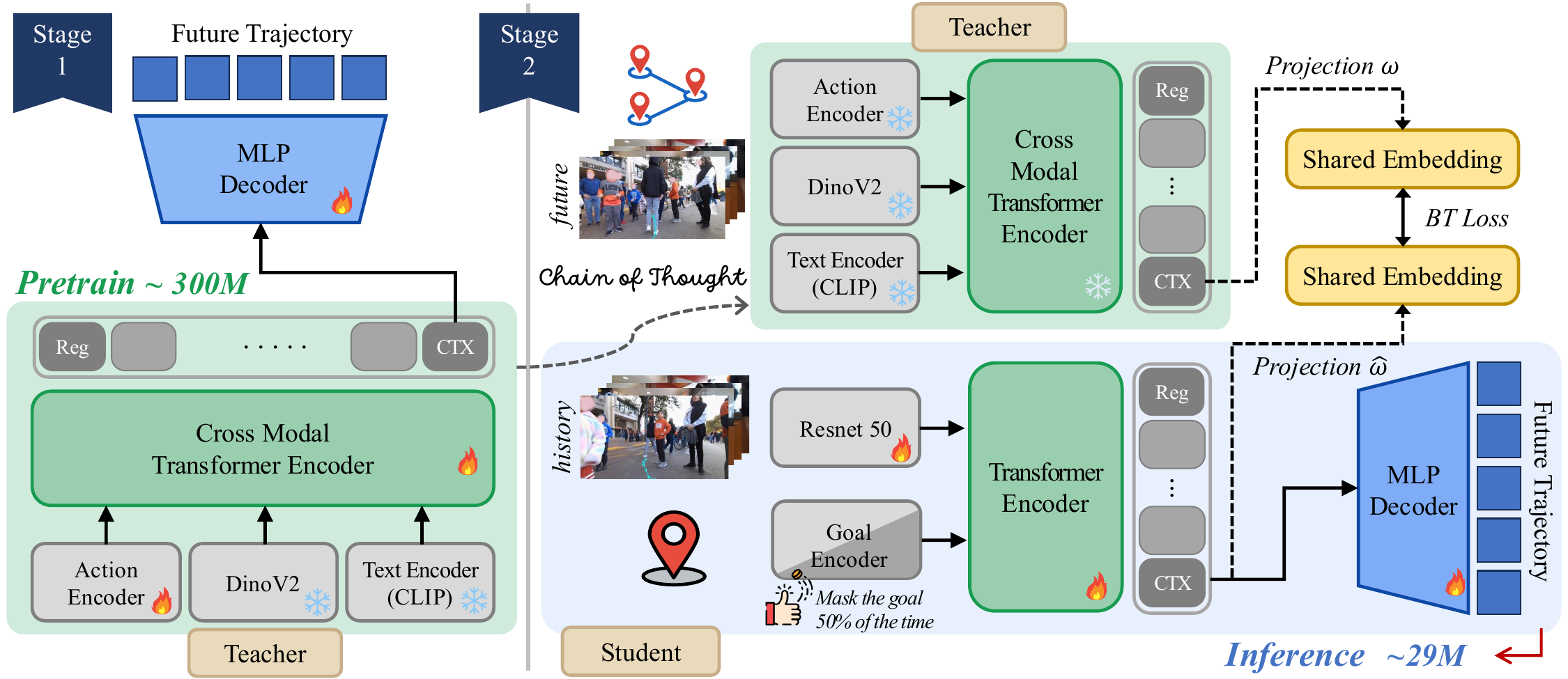}

     \caption{\narrate In Stage 1, a large Teacher Model ($\sim$330M parameters) is pre-trained to predict trajectories by leveraging multi-modal inputs—observations, low-level actions, and textual signals—to embed a comprehensive understanding for each. In Stage 2, a lightweight Student Model ($\sim$29M parameters) is trained using a redundancy-reducing Barlow Twins loss (BT Loss) to embed future state representations and implicitly incorporate human-like reasoning into current state visual representations, utilizing only RGB history and goal information. Finally, the downstream task trains the Student Decoder to generate low-level motion commands.}
     \vspace{-2em}
    \label{fig:narrate2nav}
\end{figure*}
We introduce~\textsc{Narrate2Nav}, a real-time vision-action model for robot navigation. Our approach leverages a novel training framework that integrates a unified multi-modal latent representation, learnable goal prediction, and implicit human-like language reasoning, enabling efficient navigation in dynamic, human-centric environments.

\subsection{Problem Formulation}

We formulate visual navigation as the task of driving a robot through a dynamic, human-centric environment using only RGB images from a front-facing camera, as explored in prior work~\cite{vanp,gnm}. At each time step \( t \), the robot receives a sequence of recent observations
\( o_t = \{I_{t-\tau}\}_{\tau=0}^{N-1} \in \mathcal{O} \)
comprising RGB images $I$ from the past \( N \) frames, and a goal position $g$ specified in the robot’s current 2D coordinate frame. The objective is to produce navigation action $a$ defined as a sequence of future low-level motion commands (e.g., trajectory waypoints) for short-horizon navigation. Our objective is to train a policy \( \pi_\theta(a \mid o, g) = P(a \mid o, g; \theta) \), parameterized by \( \theta \), that produces actions by conditioning on both observations and goals, generating low-level motion commands for visual navigation at each time step.

\subsection{Architecture}
The \narrate architecture comprises a two-stage training pipeline: a Teacher Model and a Student Model (see  Figure~\ref{fig:narrate2nav}). 

In Stage 1, a large Teacher Model ($\sim$330M parameters) is pre-trained to predict future trajectories by leveraging multi-modal inputs—visual history, future observations, low-level actions, and CoT textual reasoning. In Stage 2, a lightweight Student Model ($\sim$29M parameters) is trained via a redundancy reduction Barlow Twin loss (BT Loss) to implicitly embed human-like reasoning into visual representations, using only RGB history and goal information. At inference, \textsc{Narrate2Nav} operates in real time using RGB inputs, bridging visual observations to low-level motion commands while implicitly incorporating language-driven reasoning.

\subsubsection{Language as a modality}

We employ natural language to endow \textsc{Narrate2Nav} with human-like reasoning capabilities. To integrate this as a real-time module within a vision-language-action model, we treat language as a distinct modality for each state. Specifically, we generate natural language narrations describing the robot's egocentric view for each interaction, capturing its perception and predictions of surrounding human movements. These narrations serve as inputs for CoT reasoning to produce actions. Additionally, we use language as a weak signal to estimate distance measurements, replacing traditional LiDAR and 3D visual inputs with lightweight textual descriptions of spatial relationships. We also incorporate a general description of the goal position and the robot's future navigation plan. At each step, we overlay waypoints for the next 2.5 seconds (5 frames at 2~Hz) onto the RGB image and prompt a VLM to describe the trajectory in relation to human positions within the scene.
As highlighted by prior research~\cite{spatialgpt}, current SoTA VLMs exhibit limitations in spatial reasoning, a critical capability for navigation tasks. To address this, we evaluate multiple VLMs and select Qwen2-VL-72B-Instruct~\cite{qwen} for its superior performance in our context.

\subsubsection{Stage 1: Teacher Model Training}
\narrate embeds a comprehensive understanding of each state, including the contextual information of the scene and the potential future movements of nearby humans, into a large Teacher Model ($\sim$330M parameters), then to be infused into a smaller Student Model. The Teacher Model is trained in a supervised, end-to-end manner. Given the $n$ frames of context (referred to as future context in the stage 2), the next $n$ trajectories, and textual reasoning about the scene, the model is tasked with generating the next $n$ trajectories. The main goal is to learn an informative embedding for each state. Therefore, the input trajectory and the output trajectory are the same, while the input is augmented with visual and textual signals.
The Teacher Model comprises text, vision, and action encoders, whose output embeddings are used as input to a cross-modal transformer consisting of 6 layers, a hidden dimension of 256, and 8 attention heads.

\begin{wrapfigure}[13]{r}{0.50\columnwidth} 
    \centering    \includegraphics[width=0.5\columnwidth]{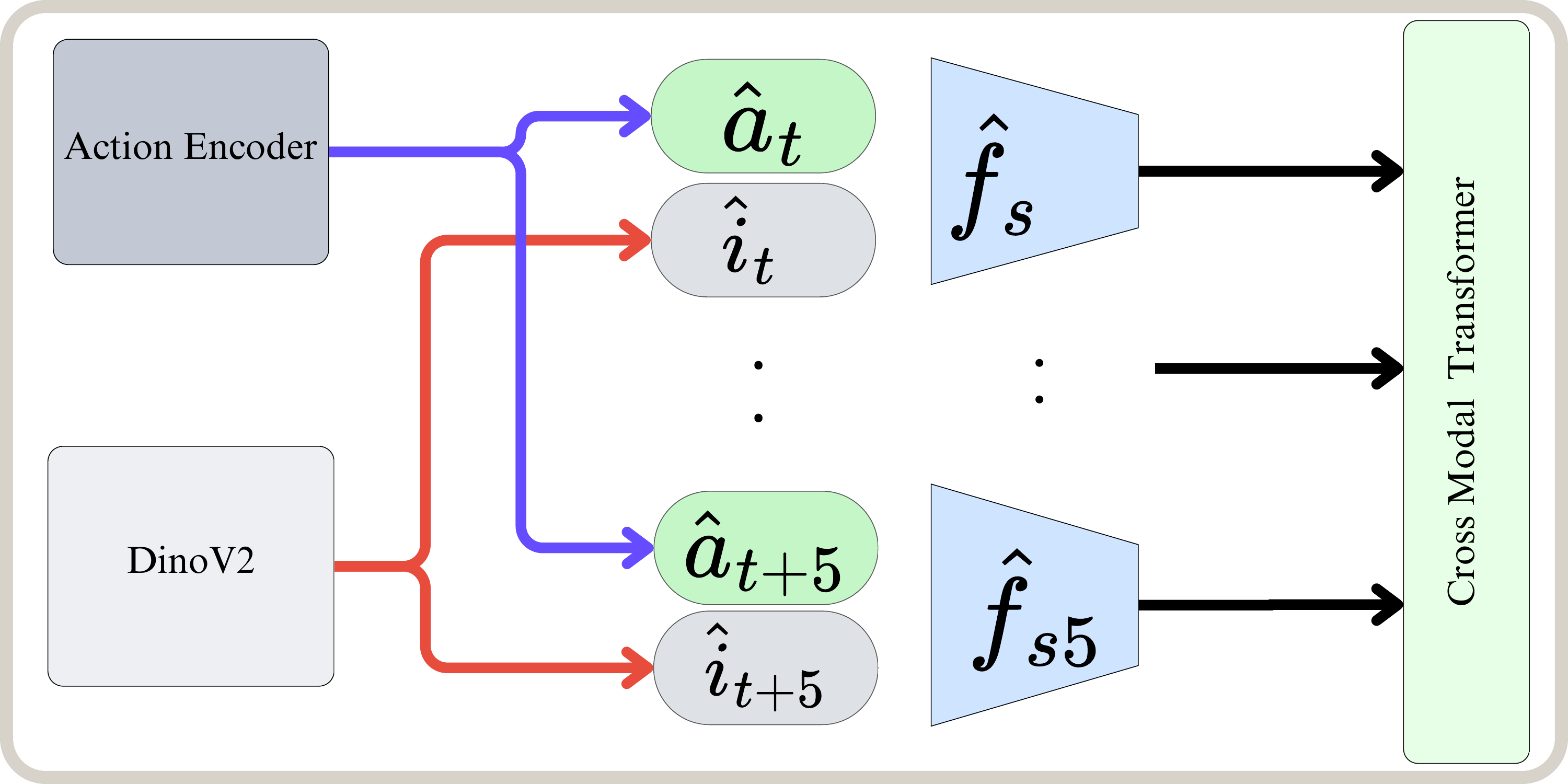} 
    \caption{Early fusion of action and observation embeddings, resulting in an effective embedding of each state.}
    \label{fig:fusion}
\end{wrapfigure}

\textnormal{\textbf{\footnotesize Unified Multi-Modal Latent Representation:}} We employ early fusion of modalities (action, vision) for each state, subsequently passing them as unified tokens to the cross-modal transformer. Specifically, at each time step, given encoded RGB observations $o_t = \text{Seq}(I_{t + \tau_f} \rightarrow I_t)$  and corresponding encoded waypoint trajectories $a_t = \text{Seq}(a_{t + \tau_f} \rightarrow a_t)~\text{where } \tau_f \in [0, n]$, we compute a unified representation of the state as: $z_t = f_s(\hat{a}_t, \hat{i}_t)$
where $f_s$ maps all embeddings to a shared latent space, mitigating statistical variance across modalities. As shown in Figure \ref{fig:fusion}, this alignment yields a uniform representation across modalities, streamlining multi-modal integration in the subsequent encoding layers. Our empirical analysis supports this architecture over interleaved ($a_{1}$, $o_{1}$, $a_{2}$, $o_{2}$, ..., $a_{n}$, $o_{n}$) or grouped ($a_{1}$, ..., $a_{n}$, $o_{1}$, ..., $o_{n}$) representation, which either do not converge or take significant computation to understand the relation of each observation to its corresponding action.

The Teacher Model uses a pretrained and frozen DINOv2~\cite{dinov2} as the vision encoder to extract visual features, while the 2D $(X, Y)$ coordinates of the waypoints are projected into the embedding space using a linear transformation $\hat{a}_k = g(a_k) = W_a \cdot [X_k, Y_k]^T + b_a$, where $a_k$ denotes the $k$-th waypoint in the trajectory sequence. For the text input, we use the pretrained CLIP~\cite{clip} text encoder $\hat{t} = f_{\text{CLIP}}(T), \ \hat{t} \in \mathbb{R}^{d_t}$, where $d_t$ is the dimension of the text embedding. All embeddings are projected to a 256-dimensional space. As the focus is on learning non-fixed, relative relationship patterns among RGB, text, and waypoints in short-length sequences (non-fixed as text does not always refer to a specific frame), we choose learned positional encodings to adaptively capture these dynamic, task-specific interactions, rather than the well-established sinusoidal encodings used in natural language processing for absolute positional information. Therefore, we encode temporal state using a learnable positional encoding $P \in \mathbb{R}^{(1 + \tau_f + 2) \times 256}$ in the input sequence. The final transformer input, concatenated with a register token and context token, is:
\[
Z_{\text{final}} = [\text{reg}, z_{t }, \dots, z_{t+\tau_f}, \hat{t}, \text{ctx}] + P, \quad Z_{\text{final}} \in \mathbb{R}^{(1 + \tau_f + 2) \times 256}.
\]
Finally, the Teacher Model (Figure \ref{fig:narrate2nav}, left model), with a fully connected MLP decoder, is trained end-to-end to predict $\text{Seq}(a_{t} \rightarrow a_{t+ \tau_f})$.

\subsubsection{Stage 2: Main (Student) Model Training}

As we define our task on RGB-only input, for the main model (only the main model is used at inference time), we pass only the sequence of RGB $o_t = \text{Seq}(I_{t + \tau_p} \rightarrow I_t) \text{ where } \tau_p \in [-n, 0]$ and one single point as the goal in 2D $(X, Y)$ coordinates. Similar to the Teacher Model the learnable positional encoding is attached to the input. The final input to the transformer is:

\[
Z_{\text{final}} = [\text{reg}, z_{t - \tau_P}, \dots, z_t, \hat{t}, \text{ctx}] + P, \quad Z_{\text{final}} \in \mathbb{R}^{(1 + \tau_P + 2) \times 256}.
\]
The Student Model uses Resnet50~\cite{resnet50} architecture as the vision encoder and encodes the 2D $(X, Y)$ goal using a linear transformation.

{\footnotesize \textbf{Goal Infilling:}} 
To develop a unified policy supporting both goal-directed navigation and undirected exploration, we use a stochastic, learnable masked modeling approach. In this framework, the goal is replaced with a learnable embedding for 50\% of the training instances. Therefore, there are two learned embeddings for the goal: one for when the goal is explicitly provided, guiding task-oriented behavior, and another for when the goal is masked, promoting task-agnostic exploration. This method enables joint training of task-agnostic and task-oriented behaviors, effectively mitigating challenges posed by the absence of explicit goals~\cite{vertiformer}.

Once the supervised end-to-end training of the Teacher Model is complete, we freeze all its modules and remove its decoder. We then distill its embeddings into the main (student) model.
Specifically, at each step of training, the inputs passed to the Teacher and Student Models are $\text{Seq}(I_{t+1} \rightarrow I_{t + \tau_f})$ and $\text{Seq}(I_{t -\tau_P} \rightarrow I_t)$ respectively, where $\tau_P$ denotes the past frames (e.g., $[I_{-5}, I_{-4}, \ldots, I_0]$) and $\tau_f$ denotes the future frames (e.g., $[I_1, I_2, \ldots, I_5]$). Typically, we set $\tau_P = \tau_f = 5$ and predict the corresponding sequence of actions for the next five time steps. \narrate maximizes the shared information between the past-to-current state and future states by applying Barlow Twins~\cite{barlow} loss on the projected context tokens ($ctx$) of the two models' transformers, thereby aligning past observations with the future multi-modal feature space. In contrast to vision SSL models that rely on joint embeddings of augmented images, \narrate aligns the future multi-modal feature space—encompassing action space $\mathcal{A}$, text space $\mathcal{T}$, and pixel latent space $\mathcal{O}$—with the latent space of past pixels. This alignment is enforced through the Barlow Twins loss, defined as:
\[
\mathcal{L}_{\text{BT}} = \sum_{i} (1 - \mathcal{C}_{ii})^2 + \lambda \sum_{i} \sum_{j \ne i} \mathcal{C}_{ij}^2
\]
where $\mathcal{C} = \frac{Z_{\text{past}}^\top Z_{\text{future}}}{\|Z_{\text{past}}\|\|Z_{\text{future}}\|}$ is the cross-correlation matrix computed between the normalized embeddings from past pixel features $Z_{\text{past}}$ and future multi-modal features $Z_{\text{future}}$, and $\lambda$ is a weighting factor that balances the invariance and redundancy reduction terms.
After completing the pretraining, we add the decoder and train the model end-to-end using Mean Squared Error (MSE) between the model’s predictions and the corresponding ground truth value.

\subsection{Implementation}
We utilize a selected subset of the SCAND~\cite{karnan2022scand} dataset, gathered from varied, densely populated public environments with intricate human-robot interaction scenarios. Additionally, we collect another dataset to ensure a robust evaluation on out-of-distribution, unseen data, providing a fair comparison across all baselines and our model. We collected data using an AgileX Scout Mini robot equipped with a ZED2 stereo camera, leveraging visual odometry to estimate the robot's position and orientation relative to its own body frame. The robot is driven at linear velocities ranging from 0 to 1.6~m/s and angular velocities ranging from $-1.5$ to 1.5~rad/s, consistent with the velocity ranges of the SCAND~\cite{karnan2022scand} dataset. The collected dataset encompasses a variety of complex social scenarios, including Navigation in Crowds, Frontal Approach, Human Following, Narrow Passageway, and Intersection.~\cite{pirk2022protocol}. 
We train our model on 8 × A100 GPUs (40 GB memory per GPU) using the AdamW optimizer with a learning rate of 2e-4 for 267 epochs in stage one, and a total (pretraining + downstream task) of 747 epochs in stage two. 
During deployment, the model takes as input the goal coordinates relative to the robot's frame, along with five past images encompassing the most recent 2.5 seconds of robot history. The output trajectory predicted by the model is tracked using a Pure Pursuit controller with a dynamic lookahead distance of at-least 0.2m in front of the robot.

\section{Analysis}
\label{sec::Analysis}

In this section, we discuss comparative evaluations, and ablation studies of our approach. 
The experiments are designed to highlight the advantages introduced by our proposed innovations. 
\subsection{Quantitative Analysis}
We evaluate~\textsc{Narrate2Nav} in five distinct, challenging social scenarios as described by Francis et al.~\cite{francis2025principles}, encompassing both indoor and outdoor environments. We compare our approach with previous SoTA baseline models, including GNM~\cite{gnm}, ViNT~\cite{vint}, NoMaD~\cite{nomad}, and CityWalker~\cite{citywalker}. We acknowledge OLiVia-Nav~\cite{vlm-social} as closest approach to our work; however, due to the unavailability of its open-source code, implementation details, and prompt design, we were unable to reproduce its results for a direct comparison. Table~\ref{tab:results-offline} demonstrates comparative performance of \narrate to other when goal is provided in unseen offline dataset.

\begin{table*}[htb]
\centering
\caption{Comparison of \narrate with four SoTA Methods on the Unseen Dataset.}
\label{tab:results-offline}

\renewcommand{\arraystretch}{1.3} 

\begingroup
\Large 
\resizebox{\textwidth}{!}{%
\begin{NiceTabular}[rules/width=1pt]{@{}l*{18}{@{\hspace{2pt}}c@{\hspace{2pt}}}@{}}

\toprule
\multirow{2}{*}{\textbf{Method}} &
\multicolumn{3}{c}{\textbf{Navigation in Crowds}} &
\multicolumn{3}{c}{\textbf{Frontal Approach}} &
\multicolumn{3}{c}{\textbf{Human Following}} &
\multicolumn{3}{c}{\textbf{Narrow Passageway}} &
\multicolumn{3}{c}{\textbf{Intersection}} &
\multicolumn{3}{c}{\textbf{All}} \\
\cmidrule(lr){2-4} \cmidrule(lr){5-7} \cmidrule(lr){8-10} \cmidrule(lr){11-13} \cmidrule(lr){14-16} \cmidrule(lr){17-19}
& \textbf{↓AOE} & \textbf{↓ADE} & \textbf{↓FDE} &
\textbf{↓AOE} & \textbf{↓ADE} & \textbf{↓FDE} &
\textbf{↓AOE} & \textbf{↓ADE} & \textbf{↓FDE} &
\textbf{↓AOE} & \textbf{↓ADE} & \textbf{↓FDE} &
\textbf{↓AOE} & \textbf{↓ADE} & \textbf{↓FDE} &
\textbf{↓AOE} & \textbf{↓ADE} & \textbf{↓FDE} \\
\midrule
GNM & 0.20 & 0.42 & 0.84&        0.17 & 0.36 & 0.72&           0.13 & 0.39 & 0.79              & 0.23 & 0.61 & 1.20&        0.19 & 0.36 & 0.74&            0.13 & 0.38 & 0.71\\

ViNT & 0.18 & 0.54 & 1.00 &    0.19 & 0.51 & 0.95&               0.17 & 0.55 & 1.03&       
   0.30 & 0.79 & 1.47&      0.20 & 0.48 & 0.94&        0.17 & 0.50 & 0.93\\

NoMaD & 0.13 & 0.40 & 0.73&    0.07 & 0.29 &  0.55&          0.04 & 0.34 & 0.61 &  
        0.09 & 0.53 & 0.93&             0.10 & 0.30 & 0.57&          
        0.09 & 0.34 & 0.61\\

CityWalker & 0.84 & 0.95 & 1.75& 0.80 & 0.90 & 1.67&             0.84 & 0.98 & 1.83& 
          0.75 & 1.08 & 2.02&            0.83 & 0.90 & 1.66&              0.87 & 0.93 & 1.71\\
\specialrule{1pt}{1pt}{0pt}

\textbf{Narrate2Nav} & \textbf{0.05} & \textbf{0.16} & \textbf{0.23} & \textbf{0.01} & \textbf{0.10} & \textbf{0.13} & \textbf{0.01} & \textbf{0.14} & \textbf{0.20} & \textbf{0.04} & \textbf{0.17} & \textbf{0.24} & \textbf{0.02} & \textbf{0.13} & \textbf{0.18} & \textbf{0.04} & \textbf{0.16} & \textbf{0.24} \\

\bottomrule
\end{NiceTabular}%
}
\endgroup
\end{table*}
\vspace{-4pt}
Note that the \textsc{All} column does not represent an average of the other columns; instead, it is an evaluation of a broader set of unlabeled samples along with the scenario-specific samples, providing a holistic assessment of prediction accuracy throughout the sequence.
\textit{Final Displacement Error} (FDE) focuses on the spatial deviation at the final time step. 
Additionally, we use the \textit{Average Orientation Error} (AOE) as defined by Liu et al.~\cite{citywalker}, which measures the angular difference between the predicted and ground truth trajectory vectors.
Our analysis shows that ~\textsc{Narrate2Nav} outperforms the baselines by 52.94\% over all the scenarios, thanks to the enriched visual embedding with social context, implicit human-like language reasoning, and pretext training method.

\subsection{Qualitative Analysis}

We present a qualitative analysis of the learned activation maps and predicted trajectories. To visualize the attention maps, we apply the Jet colormap, with red denoting areas of highest attention, on the last layer of each model's vision encoder. Since CityWalker uses a pretrained, frozen vision encoder, we do not show it here.
\setlength{\abovecaptionskip}{2pt}  
\setlength{\belowcaptionskip}{1pt}

\vspace{-8pt}
\begin{figure}[h]
    \centering
    
    \begin{subfigure}[t]{0.245\textwidth}
        \centering
        \includegraphics[width=\textwidth]{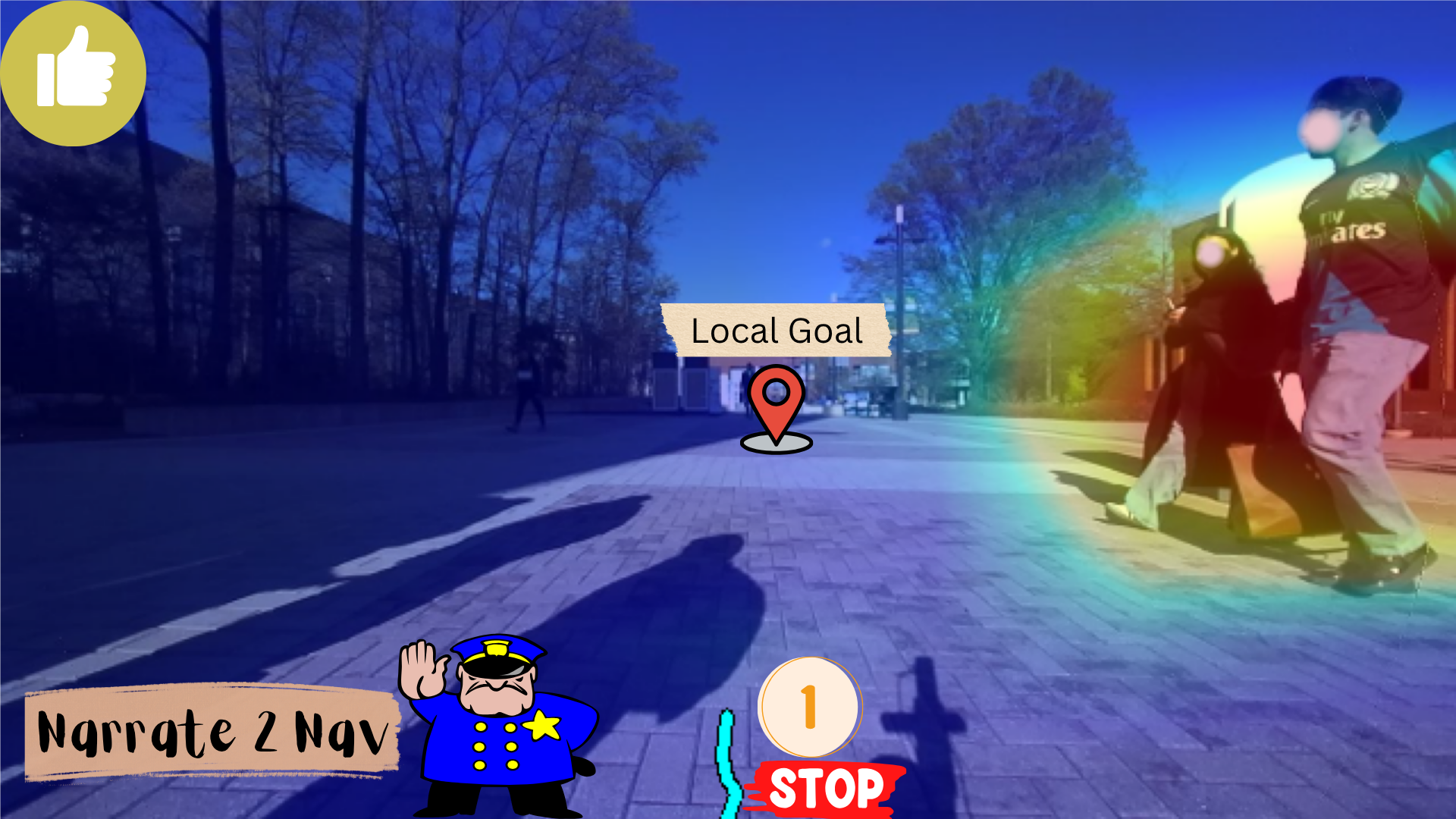} 
    \end{subfigure}
    \hfill
    \begin{subfigure}[t]{0.245\textwidth}
        \centering
        \includegraphics[width=\textwidth]{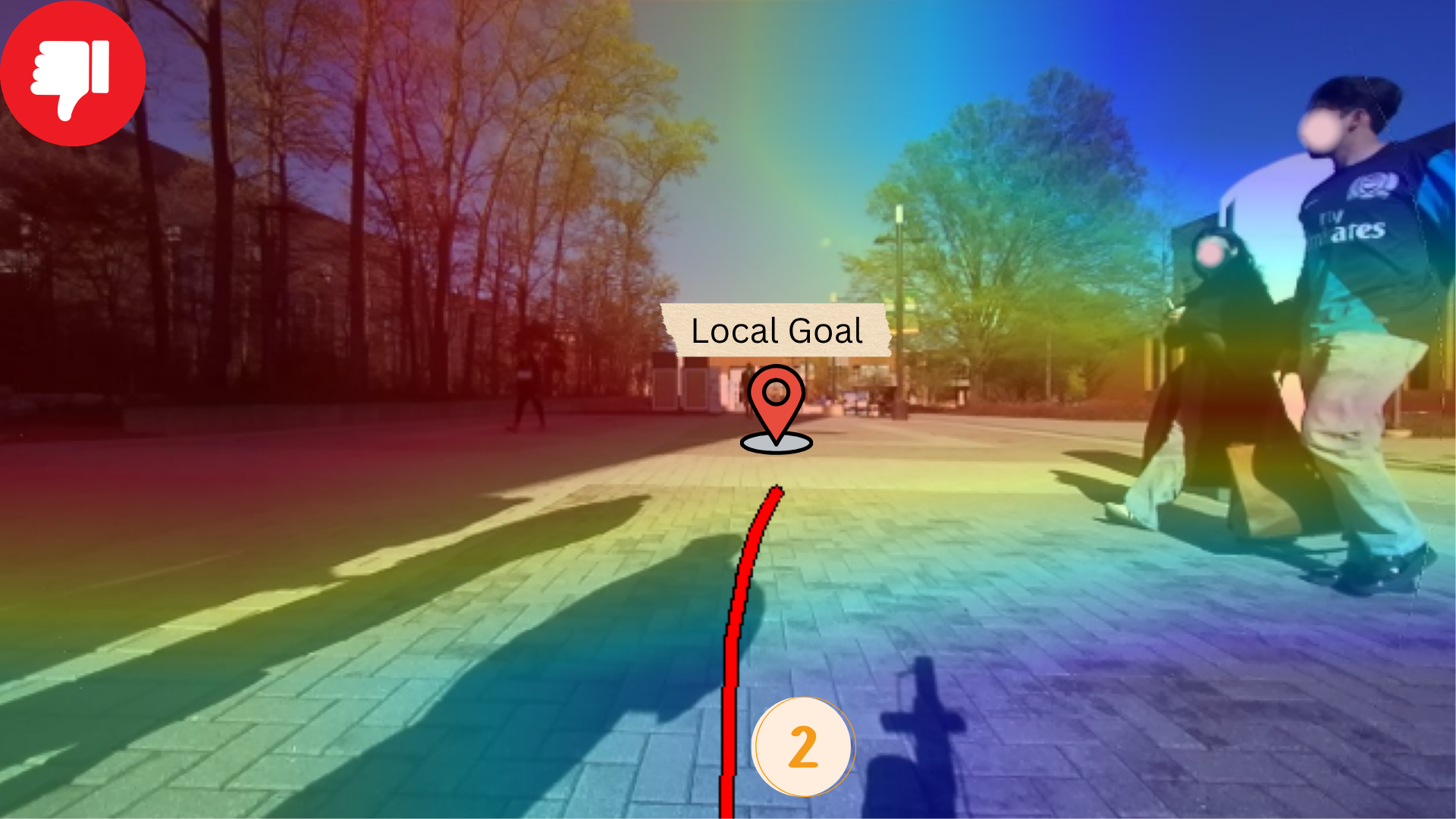} 

    \end{subfigure}
    \hfill
    \begin{subfigure}[t]{0.245\textwidth}
        \centering
        \includegraphics[width=\textwidth]{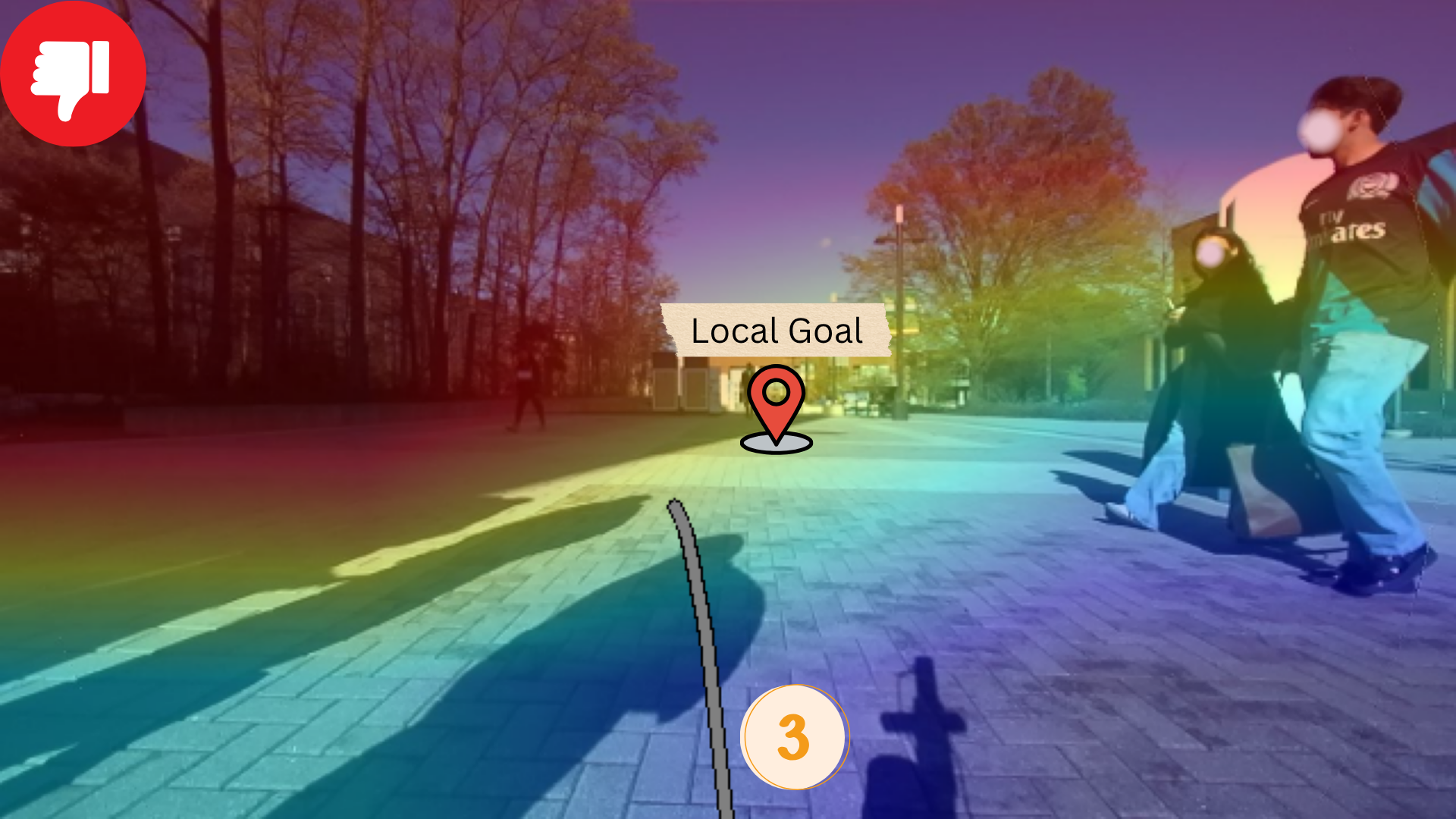} 
      
    \end{subfigure}
    \hfill
    \begin{subfigure}[t]{0.245\textwidth}
        \centering
        \includegraphics[width=\textwidth]{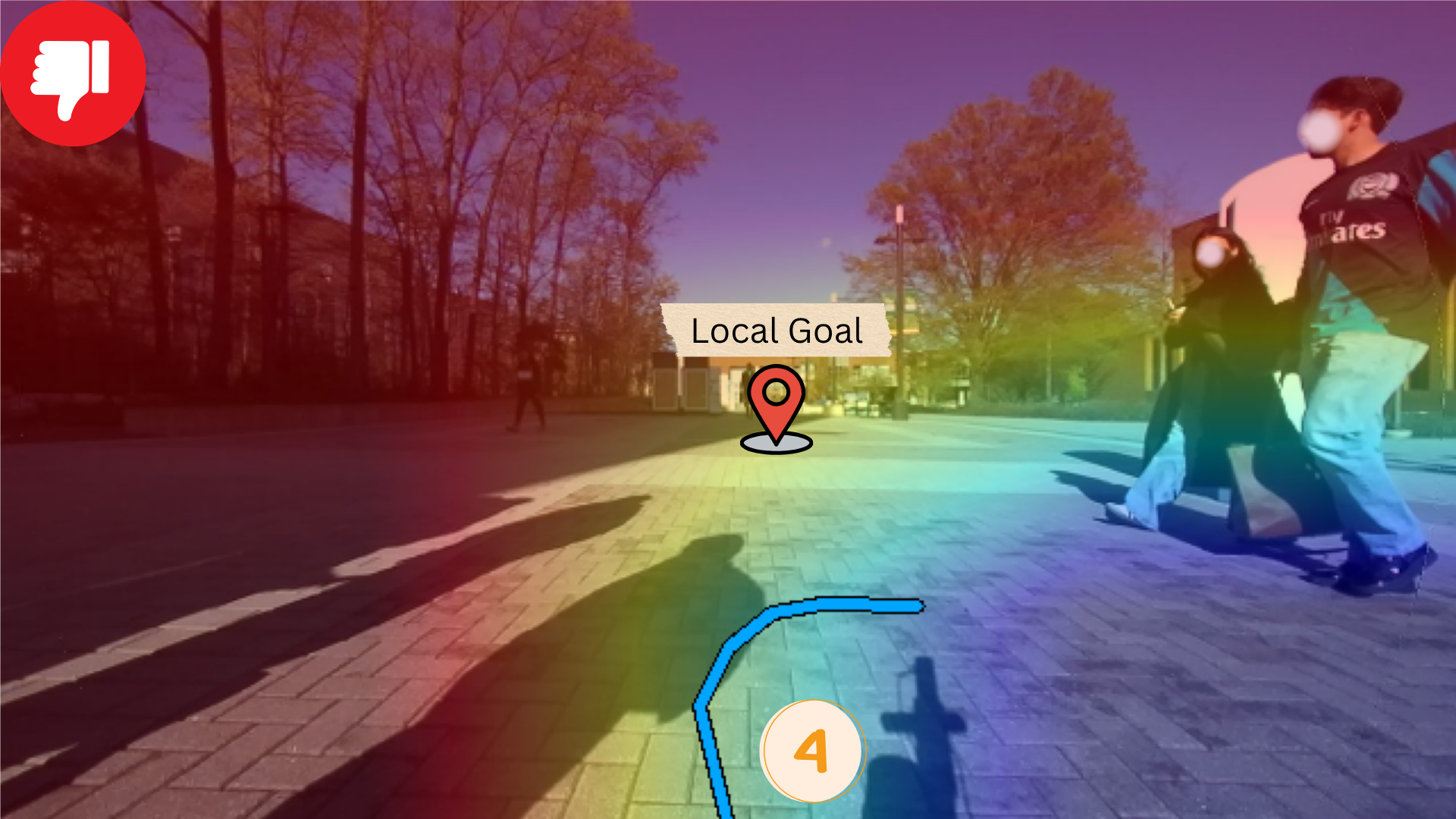} 
    \end{subfigure}
    
    \begin{subfigure}[t]{0.245\textwidth}
        \centering
        \includegraphics[width=\textwidth]{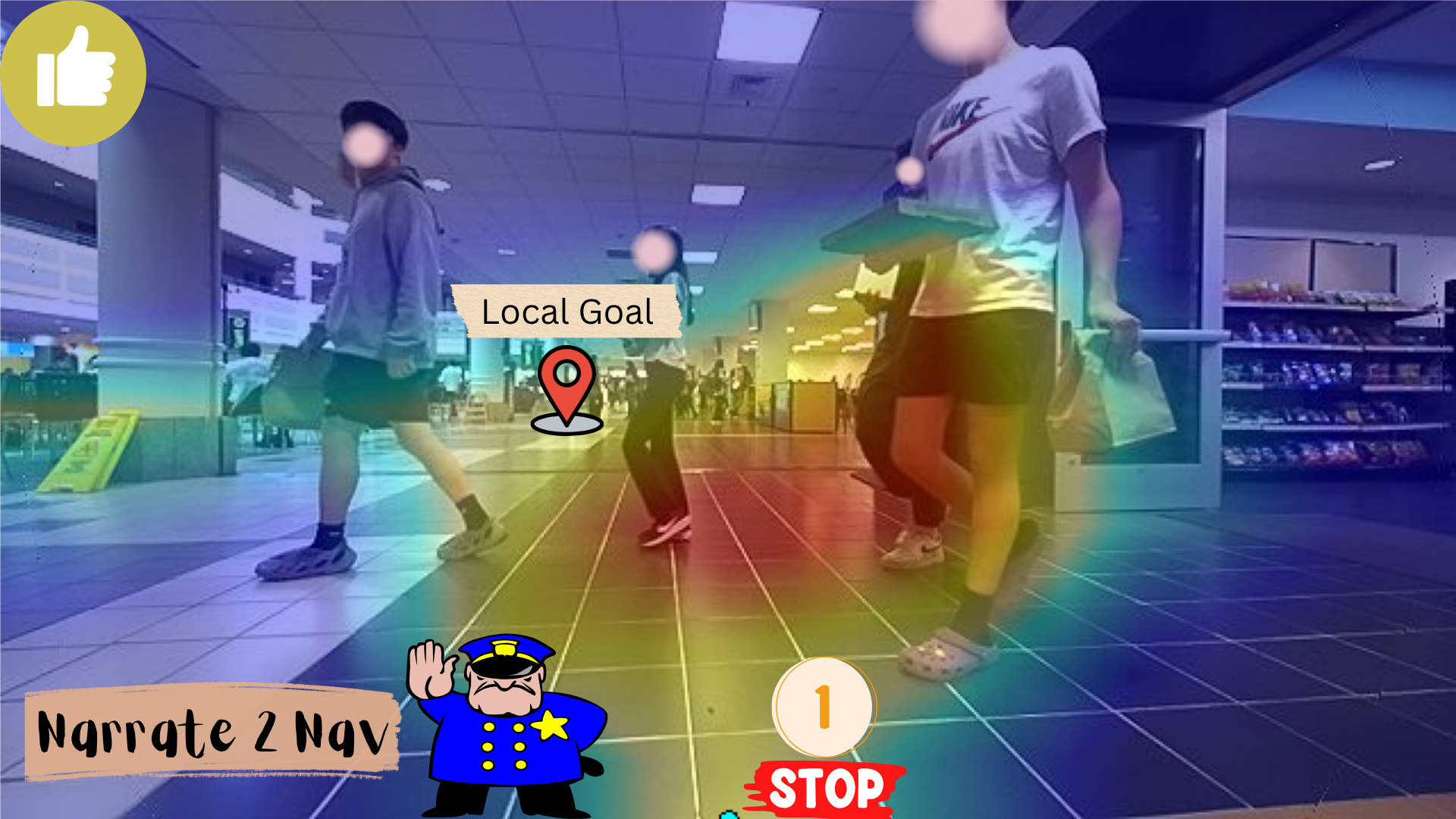} 
        \caption*{Narrate2Nav}
    \end{subfigure}
    \hfill
    \begin{subfigure}[t]{0.245\textwidth}
        \centering
        \includegraphics[width=\textwidth]{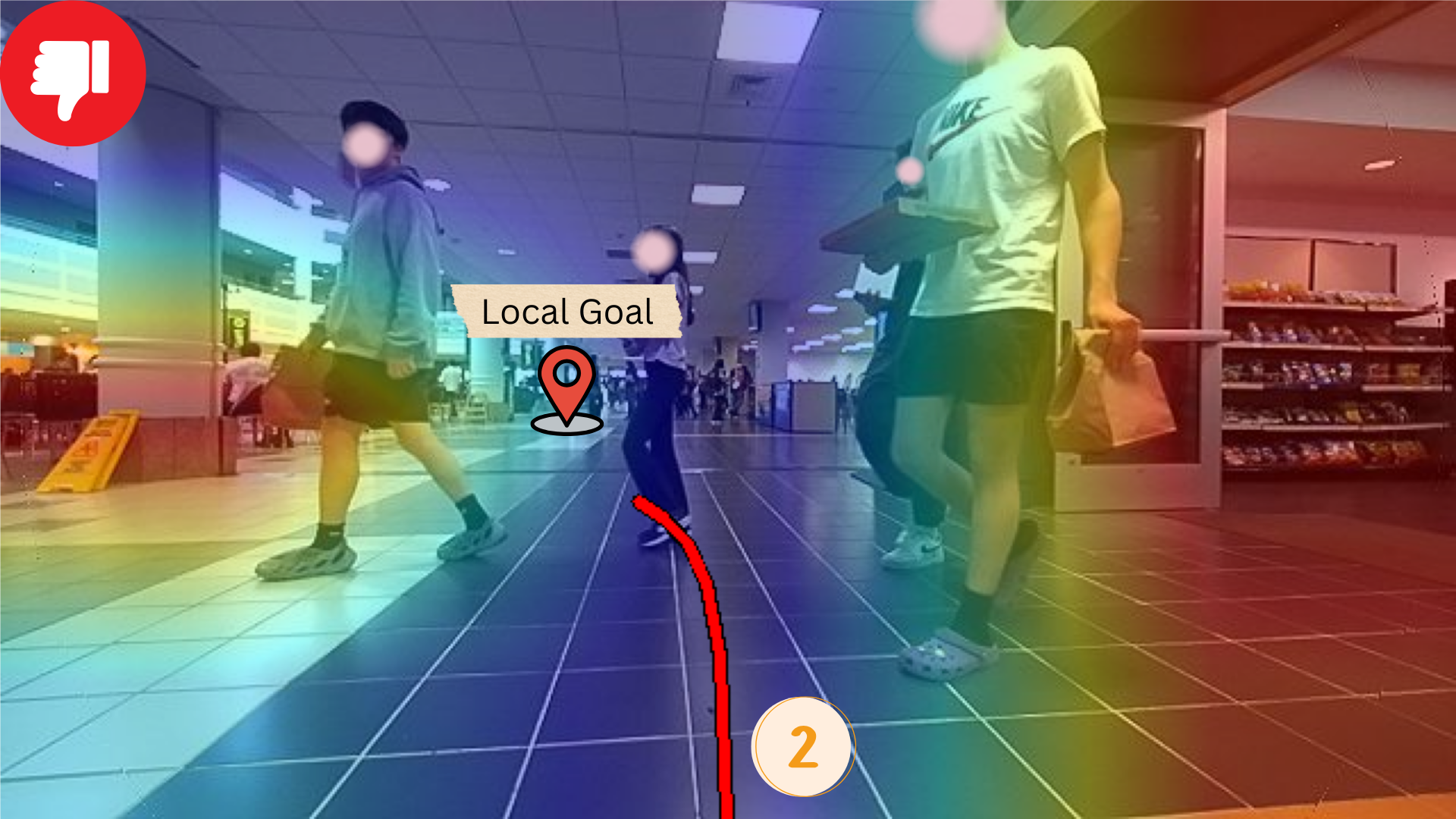} 
        \caption*{NoMAD}
    \end{subfigure}
    \hfill
    \begin{subfigure}[t]{0.245\textwidth}
        \centering
        \includegraphics[width=\textwidth]{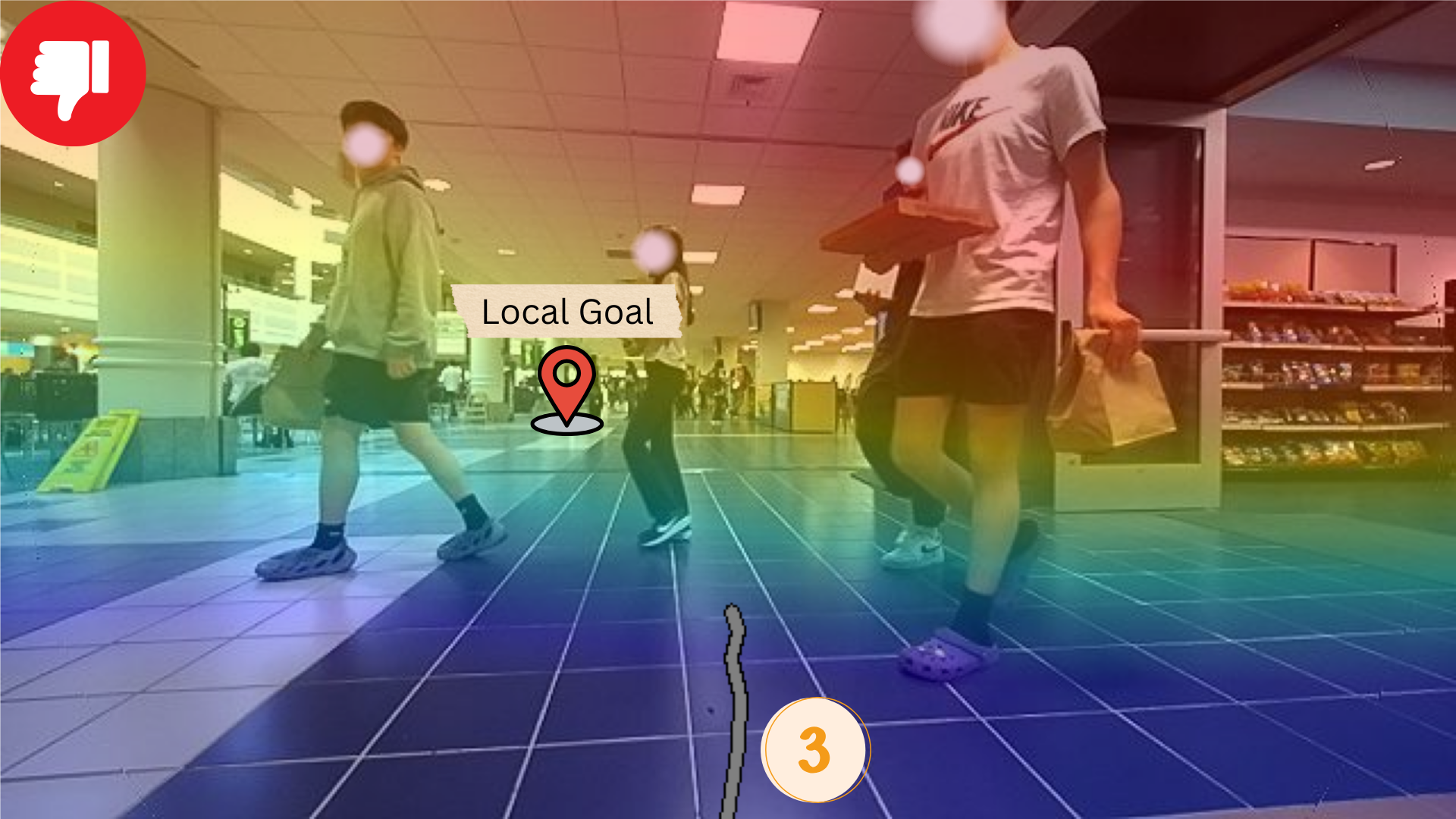} 
        \caption*{GNM}
    \end{subfigure}
    \hfill
    \begin{subfigure}[t]{0.245\textwidth}
        \centering
        \includegraphics[width=\textwidth]{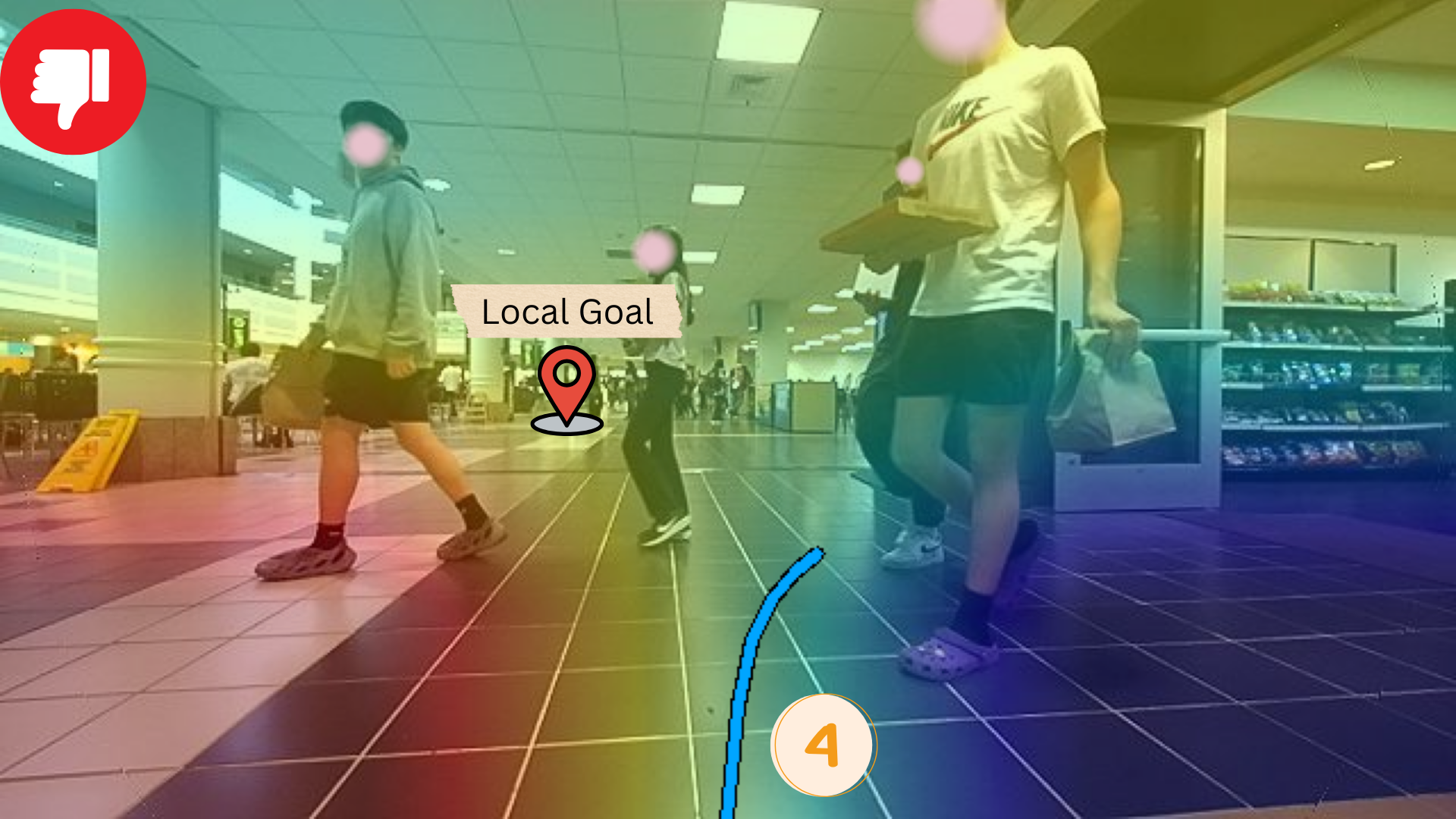} 
        \caption*{ViNT}
    \end{subfigure}
  
    \caption{Qualitative comparative analysis of activation maps and predicted trajectories (projected in 2D) in two scenarios (outdoor and indoor) in human-populated areas between \narrate and four SoTA models. 
    In both scenarios, the robot is expected to stop and yield to people, while all SoTA models generate trajectories that collide with them. \narrate's attention map clearly highlights the people obstructing its path.}
    \label{fig:attention}
\end{figure}
As shown in Figure \ref{fig:attention}, the \textsc{Narrate2Nav} attention map highlights the two people crossing its path, focusing on features relevant to navigation, while the attention maps of NoMAD~\cite{nomad}, GNM~\cite{gnm}, and ViNT~\cite{vint}(tagged as 2, 3, and 4, respectively) are less interpretable, often highlighting entire or irrelevant parts of the scene. This results in inaccurate predicted trajectories that collide with humans.
We attribute this refined attention map to the correct textual signals that directly point to regions of interest for navigation tasks.

\subsection{Real-World Analysis}

We evaluate the success rate of goal reaching and collision avoidance for our method, \narrate, across four challenging real-world scenarios, comparing its performance against baselines GNM~\cite{gnm}, ViNT~\cite{vint}, and NoMAD~\cite{nomad}. CityWalker~\cite{citywalker} was excluded from real-world deployment due to its poor performance on the offline test set. To ensure a fair assessment, we omitted one scenario, \textsc{Navigation in Crowds}, where all models, including \narrate, exhibited degraded performance. As required by their methodologies, we pre-recorded trajectory images to construct topological maps for the baselines, consistent with GNM~\cite{gnm}, ViNT~\cite{vint}, and NoMAD~\cite{nomad}. The performance results are summarized in Table~\ref{tab:results-realworld}.

\begin{table*}[htb]
\centering
\caption{Comparison of \narrate with three SoTA Methods in real-world scenarios. Collision and success rate are reported per 10 trials. \textit{'Fail'} indicates that the method failed to complete any trial in that scenario.}
\label{tab:results-realworld}

\renewcommand{\arraystretch}{1.3}
\resizebox{\textwidth}{!}{%
\begin{NiceTabular}[rules/width=1pt]{@{}l*{10}{@{\hspace{2pt}}c@{\hspace{2pt}}}@{}}
\toprule
\multirow{2}{*}{\textbf{Method}} &
\multicolumn{2}{c}{\textbf{Intersection}} &
\multicolumn{2}{c}{\textbf{Frontal Approach}} &
\multicolumn{2}{c}{\textbf{Human Following}} &
\multicolumn{2}{c}{\textbf{Narrow Passageway}} &
\multicolumn{2}{c}{\textbf{All}} \\
\cmidrule(lr){2-3} \cmidrule(lr){4-5} \cmidrule(lr){6-7} \cmidrule(lr){8-9} \cmidrule(lr){10-11}
& \textbf{↓Collision} & \textbf{↑Success Rate} &
\textbf{↓Collision} & \textbf{↑Success Rate} &
\textbf{↓Collision} & \textbf{↑Success Rate} &
\textbf{↓Collision} & \textbf{↑Success Rate} &
\textbf{↓Collision} & \textbf{↑Success Rate} \\
\midrule
GNM 
& \textbf{3/10} & \uppercase{0/10}      
& \uppercase{Fail} & \uppercase{Fail}
& 5/10 & 9/10
& \textbf{5/10} & 9/10
&4.3/10 & 6/10 \\
ViNT 
& \textbf{3/10} & 1/10
& 8/10 & 2/10
& 3/10 & \textbf{10/10}
& 7/10 & \textbf{10/10}
& 5.25/10 & 5.75/10 \\
NoMaD 
& 7/10 & 2/10
& 5/10 & 3/10
& \textbf{1/10} & \textbf{10/10}
& 7/10 & 8/10
& 5/10 & 5.75/10 \\
\specialrule{1pt}{1pt}{0pt}
\textbf{Narrate2Nav} & 
\textbf{3/10} & \textbf{10/10} & \textbf{3/10} & \textbf{10/10} & 2/10 & \textbf{10/10} & \textbf{5/10} & 4/10 & \textbf{3.25/10} & \textbf{8.5/10} \\
\bottomrule
\end{NiceTabular}%
}
\end{table*}
\vspace{-4pt}

We define the success rate as reaching the goal and a collision as any instance where the human must yield to the robot (though reaching the goal may still be possible). During experiments, changes in scene dynamics, lighting, or available space significantly impacted baseline performance, leading to lower success rates. In contrast, \narrate demonstrated robustness to these variations, maintaining consistent performance across scenarios. To address challenges in large environments, where baselines had more room to deviate from the goal, we conducted human-following experiments in narrow passageways, but this was not feasible for the intersection scenario. Notably, the narrow passageways enabled higher goal-reaching success for baselines compared to intersections and frontal approaches, as the constrained space limited deviations from the goal.

\subsection{Ablation Study}

\setlength{\belowcaptionskip}{1pt}  
\begin{wraptable}[7]{r}{0.35\columnwidth}
\centering
\footnotesize
\resizebox{0.33\columnwidth}{!}{
\begin{NiceTabular}{@{}lccc@{}}
\toprule
\multirow{2}{*}{\textbf{Method}} & \multicolumn{3}{c}{\textbf{All}} \\
\cmidrule(lr){2-4}
& \textbf{↓AOE} & \textbf{↓ADE} & \textbf{↓FDE} \\
\midrule
Ablation & 0.06 & 0.19& 0.24 \\
\textbf{Narrate2Nav} & \textbf{0.04} & \textbf{0.16} & 0.24 \\
\bottomrule
\end{NiceTabular}
}
\caption{Results on ablating the text module.}\label{tab:results-ablation}
\end{wraptable}

Table~\ref{tab:results-ablation} examines the effectiveness of incorporating social context and predictive textual signals in the model's pretext training. We ablate the textual input, following the same design shown in Figure \ref{fig:narrate2nav}, and remove only the text encoder model. Our results indicate that the overall performance of our model on the same task drops  14.8\%. This clearly show the effect of the text signals in our model.

\section{CONCLUSION}
\label{sec::conclusions}
\vspace{-3pt}
In this work, we propose \textsc{Narrate2Nav}, a self-supervised learning-based approach to real-time vision-action modeling that integrates human-like language reasoning for context-aware navigation in human-centric environments. Using a novel pretraining method with Barlow Twins loss and multi-modal future state representations, it embeds social and contextual cues into a visual encoder, enabling efficient RGB-only navigation. \textsc{Narrate2Nav} achieves a 52.94\% improvement in ADE and a 41.67\% improvement in real-world experiments over SoTA baselines across various challenging scenarios. Furthermore, our analysis of visual attention maps shows that \textsc{Narrate2Nav} highlights key social cues and critical regions for navigation, showing the effectiveness of our framework in complex environments. Looking ahead, our approach opens promising new avenues for integrating richer forms of human-like reasoning into real-time robotic systems. 

\section{Limitations}
\label{sec::discussion}

Our primary motivation in this work is to demonstrate the potential of textual input as a distinct modality and to propose a novel Vision-Language-Action (VLA) architecture. This architecture not only addresses the computational challenges of VLMs but also achieves data efficiency and outperforms models trained on internet-scale data.
We acknowledge the open challenges that limit our work. Despite extensive research on grounding language in pixel values, VLMs still struggle with basic spatial reasoning tasks, such as left-right differentiation, particularly in complex scenarios. Our analysis of a generated dataset reveals instances of inaccurate descriptions. These inaccuracies, combined with the challenge of determining which textual information provides a useful signal for generating improved trajectories, constrain the scalability of our approach. In this work, we primarily adopted the methodology proposed by Social-LLaVA as a signal for the vision encoder. However, there remains significant scope for research into diverse information styles to analyze how they can enhance robotic observation, next-state prediction, and subsequent action prediction. 
Given our contribution to mitigating the need for large VLMs that are not optimized for real-time inference, it is valuable to compare the performance of ~\textsc{Narrate2Nave} architecture with frameworks that directly utilize off-the-shelf large VLMs for visual robot navigation in human-populated environments. 

We identify a potential direction for future research is to benchmark our model's performance against such frameworks in real-time navigation tasks. Finally, we acknowledge that our attempt is to transfer reasoning-like features via latent embedding supervision. Further testing is required to fully verify emergent reasoning behaviors.

We recognize that our model is not yet ready for different robot embodiments or strongly robust across diverse scenarios. General microengineering, hyperparameter tuning, scenario selection, context, and, most importantly, the training data play pivotal roles in this regard. In the real-world section, we present a proof of concept demonstrating our work’s advantages over SoTA models, with the goal of motivating further research into using linguistic signals as a separate modality for improving short- and long-horizon path planning in visual navigation. The real-world experiment is subject to various factors, such as training data, human behavior, and test environment, which may affect outcomes. We strive to mitigate these limitations to deliver robust, reliable, and fair analysis, fostering further research in this field. 


\clearpage

\bibliography{root}

\begin{thebibliography}{36}
\providecommand{\natexlab}[1]{#1}
\providecommand{\url}[1]{\texttt{#1}}
\expandafter\ifx\csname urlstyle\endcsname\relax
  \providecommand{\doi}[1]{doi: #1}\else
  \providecommand{\doi}{doi: \begingroup \urlstyle{rm}\Url}\fi

\bibitem[Francis et~al.(2025)Francis, P{\'e}rez-d’Arpino, Li, Xia, Alahi, Alami, Bera, Biswas, Biswas, Chandra, et~al.]{francis2025principles}
A.~Francis, C.~P{\'e}rez-d’Arpino, C.~Li, F.~Xia, A.~Alahi, R.~Alami, A.~Bera, A.~Biswas, J.~Biswas, R.~Chandra, et~al.
\newblock Principles and guidelines for evaluating social robot navigation algorithms.
\newblock \emph{ACM Transactions on Human-Robot Interaction}, 14\penalty0 (2):\penalty0 1--65, 2025.

\bibitem[Mirsky et~al.(2024)Mirsky, Xiao, Hart, and Stone]{mirsky2024conflict}
R.~Mirsky, X.~Xiao, J.~Hart, and P.~Stone.
\newblock Conflict avoidance in social navigation—a survey.
\newblock \emph{ACM Transactions on Human-Robot Interaction}, 13\penalty0 (1):\penalty0 1--36, 2024.

\bibitem[Mavrogiannis et~al.(2023)Mavrogiannis, Baldini, Wang, Zhao, Trautman, Steinfeld, and Oh]{mavrogiannis2023core}
C.~Mavrogiannis, F.~Baldini, A.~Wang, D.~Zhao, P.~Trautman, A.~Steinfeld, and J.~Oh.
\newblock Core challenges of social robot navigation: A survey.
\newblock \emph{ACM Transactions on Human-Robot Interaction}, 12\penalty0 (3):\penalty0 1--39, 2023.

\bibitem[Xiao et~al.(2022)Xiao, Liu, Warnell, and Stone]{xiao2022motion}
X.~Xiao, B.~Liu, G.~Warnell, and P.~Stone.
\newblock Motion planning and control for mobile robot navigation using machine learning: a survey.
\newblock \emph{Autonomous Robots}, 46\penalty0 (5):\penalty0 569--597, 2022.

\bibitem[Payandeh et~al.(2023)Payandeh, Baghaei, Fayyazsanavi, Ramezani, Chen, and Rahimi]{10323465}
A.~Payandeh, K.~T. Baghaei, P.~Fayyazsanavi, S.~B. Ramezani, Z.~Chen, and S.~Rahimi.
\newblock Deep representation learning: Fundamentals, technologies, applications, and open challenges.
\newblock \emph{IEEE Access}, 11:\penalty0 137621--137659, 2023.
\newblock \doi{10.1109/ACCESS.2023.3335196}.

\bibitem[Nazeri and Bohlouli(2021)]{9679193}
M.~H. Nazeri and M.~Bohlouli.
\newblock Exploring reflective limitation of behavior cloning in autonomous vehicles.
\newblock In \emph{2021 IEEE International Conference on Data Mining (ICDM)}, pages 1252--1257, 2021.
\newblock \doi{10.1109/ICDM51629.2021.00153}.

\bibitem[Nguyen et~al.(2023)Nguyen, Nazeri, Payandeh, Datar, and Xiao]{musohu}
D.~M. Nguyen, M.~Nazeri, A.~Payandeh, A.~Datar, and X.~Xiao.
\newblock Toward human-like social robot navigation: A large-scale, multi-modal, social human navigation dataset.
\newblock In \emph{2023 IEEE/RSJ International Conference on Intelligent Robots and Systems (IROS)}, pages 7442--7447, 2023.
\newblock \doi{10.1109/IROS55552.2023.10342447}.

\bibitem[Liang et~al.(2024)Liang, Payandeh, Song, Xiao, and Manocha]{dtg}
J.~Liang, A.~Payandeh, D.~Song, X.~Xiao, and D.~Manocha.
\newblock Dtg : Diffusion-based trajectory generation for mapless global navigation.
\newblock In \emph{2024 IEEE/RSJ International Conference on Intelligent Robots and Systems (IROS)}, pages 5340--5347, 2024.
\newblock \doi{10.1109/IROS58592.2024.10802055}.

\bibitem[Raj et~al.(2024)Raj, Hu, Karnan, Chandra, Payandeh, Mao, Stone, Biswas, and Xiao]{rethinking}
A.~H. Raj, Z.~Hu, H.~Karnan, R.~Chandra, A.~Payandeh, L.~Mao, P.~Stone, J.~Biswas, and X.~Xiao.
\newblock Rethinking social robot navigation: Leveraging the best of two worlds.
\newblock In \emph{2024 IEEE International Conference on Robotics and Automation (ICRA)}, pages 16330--16337, 2024.
\newblock \doi{10.1109/ICRA57147.2024.10611710}.

\bibitem[Kretzschmar et~al.(2016)Kretzschmar, Spies, Sprunk, and Burgard]{kretzschmar16ijrr}
H.~Kretzschmar, M.~Spies, C.~Sprunk, and W.~Burgard.
\newblock Socially compliant mobile robot navigation via inverse reinforcement learning.
\newblock \emph{The International Journal of Robotics Research}, 2016.
\newblock \doi{10.1177/0278364915619772}.

\bibitem[Song et~al.(2025)Song, Liang, Payandeh, Raj, Xiao, and Manocha]{vlm-social}
D.~Song, J.~Liang, A.~Payandeh, A.~H. Raj, X.~Xiao, and D.~Manocha.
\newblock Vlm-social-nav: Socially aware robot navigation through scoring using vision-language models.
\newblock \emph{IEEE Robotics and Automation Letters}, 10\penalty0 (1):\penalty0 508--515, 2025.
\newblock \doi{10.1109/LRA.2024.3511409}.

\bibitem[Payandeh et~al.(2024)Payandeh, Song, Nazeri, Liang, Mukherjee, Raj, Kong, Manocha, and Xiao]{social-llava}
A.~Payandeh, D.~Song, M.~Nazeri, J.~Liang, P.~Mukherjee, A.~H. Raj, Y.~Kong, D.~Manocha, and X.~Xiao.
\newblock Social-llava: Enhancing robot navigation through human-language reasoning in social spaces, 2024.
\newblock URL \url{https://arxiv.org/abs/2501.09024}.

\bibitem[Narasimhan et~al.(2025)Narasimhan, Tan, Choi, and Nejat]{olivianav}
S.~Narasimhan, A.~H. Tan, D.~Choi, and G.~Nejat.
\newblock Olivia-nav: An online lifelong vision language approach for mobile robot social navigation, 2025.
\newblock URL \url{https://arxiv.org/abs/2409.13675}.

\bibitem[Payandeh et~al.(2024)Payandeh, Pluth, Hosier, Xiao, and Gurbani]{susceptible}
A.~Payandeh, D.~Pluth, J.~Hosier, X.~Xiao, and V.~K. Gurbani.
\newblock How susceptible are {LLM}s to logical fallacies?
\newblock In N.~Calzolari, M.-Y. Kan, V.~Hoste, A.~Lenci, S.~Sakti, and N.~Xue, editors, \emph{Proceedings of the 2024 Joint International Conference on Computational Linguistics, Language Resources and Evaluation (LREC-COLING 2024)}, pages 8276--8286, Torino, Italia, May 2024. ELRA and ICCL.
\newblock URL \url{https://aclanthology.org/2024.lrec-main.726/}.

\bibitem[Song et~al.(2025)Song, Liang, Xiao, and Manocha]{song2025vl}
D.~Song, J.~Liang, X.~Xiao, and D.~Manocha.
\newblock Vl-tgs: Trajectory generation and selection using vision language models in mapless outdoor environments.
\newblock \emph{IEEE Robotics and Automation Letters}, 10\penalty0 (6):\penalty0 5791--5798, 2025.
\newblock \doi{10.1109/LRA.2025.3559822}.

\bibitem[Zbontar et~al.(2021)Zbontar, Jing, Misra, LeCun, and Deny]{barlow}
J.~Zbontar, L.~Jing, I.~Misra, Y.~LeCun, and S.~Deny.
\newblock Barlow twins: Self-supervised learning via redundancy reduction.
\newblock In \emph{International conference on machine learning}, pages 12310--12320. PMLR, 2021.

\bibitem[Shah et~al.(2023{\natexlab{a}})Shah, Sridhar, Bhorkar, Hirose, and Levine]{gnm}
D.~Shah, A.~Sridhar, A.~Bhorkar, N.~Hirose, and S.~Levine.
\newblock Gnm: A general navigation model to drive any robot, 2023{\natexlab{a}}.
\newblock URL \url{https://arxiv.org/abs/2210.03370}.

\bibitem[Shah et~al.(2023{\natexlab{b}})Shah, Sridhar, Dashora, Stachowicz, Black, Hirose, and Levine]{vint}
D.~Shah, A.~Sridhar, N.~Dashora, K.~Stachowicz, K.~Black, N.~Hirose, and S.~Levine.
\newblock Vint: A foundation model for visual navigation, 2023{\natexlab{b}}.
\newblock URL \url{https://arxiv.org/abs/2306.14846}.

\bibitem[Sridhar et~al.(2023)Sridhar, Shah, Glossop, and Levine]{nomad}
A.~Sridhar, D.~Shah, C.~Glossop, and S.~Levine.
\newblock Nomad: Goal masked diffusion policies for navigation and exploration, 2023.
\newblock URL \url{https://arxiv.org/abs/2310.07896}.

\bibitem[Liu et~al.(2024)Liu, Li, Jiang, Sujay, Yang, Zhang, Abanes, Zhang, and Feng]{citywalker}
X.~Liu, J.~Li, Y.~Jiang, N.~Sujay, Z.~Yang, J.~Zhang, J.~Abanes, J.~Zhang, and C.~Feng.
\newblock Citywalker: Learning embodied urban navigation from web-scale videos.
\newblock \emph{arXiv preprint arXiv:2411.17820}, 2024.

\bibitem[Eftekhar et~al.(2024)Eftekhar, Zeng, Duan, Farhadi, Kembhavi, and Krishna]{selective}
A.~Eftekhar, K.-H. Zeng, J.~Duan, A.~Farhadi, A.~Kembhavi, and R.~Krishna.
\newblock Selective visual representations improve convergence and generalization for embodied ai, 2024.
\newblock URL \url{https://arxiv.org/abs/2311.04193}.

\bibitem[Nazeri et~al.(2024)Nazeri, Wang, Payandeh, and Xiao]{vanp}
M.~Nazeri, J.~Wang, A.~Payandeh, and X.~Xiao.
\newblock Vanp: Learning where to see for navigation with self-supervised vision-action pre-training.
\newblock In \emph{2024 IEEE/RSJ International Conference on Intelligent Robots and Systems (IROS)}, pages 2741--2746. IEEE, 2024.

\bibitem[Elnoor et~al.(2025)Elnoor, Weerakoon, Seneviratne, Liang, Rajagopal, and Manocha]{vilad}
M.~Elnoor, K.~Weerakoon, G.~Seneviratne, J.~Liang, V.~Rajagopal, and D.~Manocha.
\newblock Vi-lad: Vision-language attention distillation for socially-aware robot navigation in dynamic environments, 2025.
\newblock URL \url{https://arxiv.org/abs/2503.09820}.

\bibitem[Pokhrel et~al.(2024)Pokhrel, Nazeri, Datar, and Xiao]{cahsor}
A.~Pokhrel, M.~Nazeri, A.~Datar, and X.~Xiao.
\newblock Cahsor: Competence-aware high-speed off-road ground navigation in $\mathbb {SE}(3)$.
\newblock \emph{IEEE Robotics and Automation Letters}, 9\penalty0 (11):\penalty0 9653--9660, 2024.
\newblock \doi{10.1109/LRA.2024.3457369}.

\bibitem[Roth et~al.(2024)Roth, Nubert, Yang, Mittal, and Hutter]{viplanner}
P.~Roth, J.~Nubert, F.~Yang, M.~Mittal, and M.~Hutter.
\newblock Viplanner: Visual semantic imperative learning for local navigation, 2024.
\newblock URL \url{https://arxiv.org/abs/2310.00982}.

\bibitem[Wang et~al.(2024)Wang, Tan, and Nejat]{navformer}
H.~Wang, A.~H. Tan, and G.~Nejat.
\newblock {{NavFormer}}: {{A Transformer Architecture}} for {{Robot Target-Driven Navigation}} in {{Unknown}} and {{Dynamic Environments}}.
\newblock \emph{arXiv preprint arXiv:2402.06838}, 2024.

\bibitem[Hirose et~al.(2024)Hirose, Glossop, Sridhar, Shah, Mees, and Levine]{lelan}
N.~Hirose, C.~Glossop, A.~Sridhar, D.~Shah, O.~Mees, and S.~Levine.
\newblock Lelan: Learning a language-conditioned navigation policy from in-the-wild video.
\newblock In \emph{Conference on Robot Learning}, 2024.

\bibitem[Eftekhar et~al.(2024)Eftekhar, Weihs, Hendrix, Caglar, Salvador, Herrasti, Han, VanderBil, Kembhavi, Farhadi, Krishna, Ehsani, and Zeng]{onering}
A.~Eftekhar, L.~Weihs, R.~Hendrix, E.~Caglar, J.~Salvador, A.~Herrasti, W.~Han, E.~VanderBil, A.~Kembhavi, A.~Farhadi, R.~Krishna, K.~Ehsani, and K.-H. Zeng.
\newblock The one ring: a robotic indoor navigation generalist, 2024.
\newblock URL \url{https://arxiv.org/abs/2412.14401}.

\bibitem[Cheng et~al.(2024)Cheng, Yin, Fu, Guo, Yang, Kautz, Wang, and Liu]{spatialgpt}
A.-C. Cheng, H.~Yin, Y.~Fu, Q.~Guo, R.~Yang, J.~Kautz, X.~Wang, and S.~Liu.
\newblock Spatialrgpt: Grounded spatial reasoning in vision language models, 2024.
\newblock URL \url{https://arxiv.org/abs/2406.01584}.

\bibitem[Yang et~al.(2024)Yang, Yang, Hui, Zheng, Yu, Zhou, Li, Li, Liu, Huang, Dong, Wei, Lin, Tang, Wang, Yang, Tu, Zhang, Ma, Yang, Xu, Zhou, Bai, He, Lin, Dang, Lu, Chen, Yang, Li, Xue, Ni, Zhang, Wang, Peng, Men, Gao, Lin, Wang, Bai, Tan, Zhu, Li, Liu, Ge, Deng, Zhou, Ren, Zhang, Wei, Ren, Liu, Fan, Yao, Zhang, Wan, Chu, Liu, Cui, Zhang, Guo, and Fan]{qwen}
A.~Yang, B.~Yang, B.~Hui, B.~Zheng, B.~Yu, C.~Zhou, C.~Li, C.~Li, D.~Liu, F.~Huang, G.~Dong, H.~Wei, H.~Lin, J.~Tang, J.~Wang, J.~Yang, J.~Tu, J.~Zhang, J.~Ma, J.~Yang, J.~Xu, J.~Zhou, J.~Bai, J.~He, J.~Lin, K.~Dang, K.~Lu, K.~Chen, K.~Yang, M.~Li, M.~Xue, N.~Ni, P.~Zhang, P.~Wang, R.~Peng, R.~Men, R.~Gao, R.~Lin, S.~Wang, S.~Bai, S.~Tan, T.~Zhu, T.~Li, T.~Liu, W.~Ge, X.~Deng, X.~Zhou, X.~Ren, X.~Zhang, X.~Wei, X.~Ren, X.~Liu, Y.~Fan, Y.~Yao, Y.~Zhang, Y.~Wan, Y.~Chu, Y.~Liu, Z.~Cui, Z.~Zhang, Z.~Guo, and Z.~Fan.
\newblock Qwen2 technical report, 2024.
\newblock URL \url{https://arxiv.org/abs/2407.10671}.

\bibitem[Oquab et~al.(2024)Oquab, Darcet, Moutakanni, Vo, Szafraniec, Khalidov, Fernandez, Haziza, Massa, El-Nouby, Assran, Ballas, Galuba, Howes, Huang, Li, Misra, Rabbat, Sharma, Synnaeve, Xu, Jegou, Mairal, Labatut, Joulin, and Bojanowski]{dinov2}
M.~Oquab, T.~Darcet, T.~Moutakanni, H.~Vo, M.~Szafraniec, V.~Khalidov, P.~Fernandez, D.~Haziza, F.~Massa, A.~El-Nouby, M.~Assran, N.~Ballas, W.~Galuba, R.~Howes, P.-Y. Huang, S.-W. Li, I.~Misra, M.~Rabbat, V.~Sharma, G.~Synnaeve, H.~Xu, H.~Jegou, J.~Mairal, P.~Labatut, A.~Joulin, and P.~Bojanowski.
\newblock Dinov2: Learning robust visual features without supervision, 2024.
\newblock URL \url{https://arxiv.org/abs/2304.07193}.

\bibitem[Radford et~al.(2021)Radford, Kim, Hallacy, Ramesh, Goh, Agarwal, Sastry, Askell, Mishkin, Clark, et~al.]{clip}
A.~Radford, J.~W. Kim, C.~Hallacy, A.~Ramesh, G.~Goh, S.~Agarwal, G.~Sastry, A.~Askell, P.~Mishkin, J.~Clark, et~al.
\newblock Learning transferable visual models from natural language supervision.
\newblock In \emph{International conference on machine learning}, pages 8748--8763. PmLR, 2021.

\bibitem[He et~al.(2015)He, Zhang, Ren, and Sun]{resnet50}
K.~He, X.~Zhang, S.~Ren, and J.~Sun.
\newblock Deep residual learning for image recognition, 2015.
\newblock URL \url{https://arxiv.org/abs/1512.03385}.

\bibitem[Nazeri et~al.(2025)Nazeri, Pokhrel, Card, Datar, Warnell, and Xiao]{vertiformer}
M.~Nazeri, A.~Pokhrel, A.~Card, A.~Datar, G.~Warnell, and X.~Xiao.
\newblock Vertiformer: A data-efficient multi-task transformer for off-road robot mobility, 2025.
\newblock URL \url{https://arxiv.org/abs/2502.00543}.

\bibitem[Karnan et~al.(2022)Karnan, Nair, Xiao, Warnell, Pirk, Toshev, Hart, Biswas, and Stone]{karnan2022scand}
H.~Karnan, A.~Nair, X.~Xiao, G.~Warnell, S.~Pirk, A.~Toshev, J.~Hart, J.~Biswas, and P.~Stone.
\newblock Socially compliant navigation dataset (scand): A large-scale dataset of demonstrations for social navigation.
\newblock \emph{IEEE Robotics and Automation Letters}, 2022.

\bibitem[Pirk et~al.(2022)Pirk, Lee, Xiao, Takayama, Francis, and Toshev]{pirk2022protocol}
S.~Pirk, E.~Lee, X.~Xiao, L.~Takayama, A.~Francis, and A.~Toshev.
\newblock A protocol for validating social navigation policies.
\newblock \emph{arXiv preprint arXiv:2204.05443}, 2022.

\end{thebibliography}

\clearpage
\end{document}